\documentclass[conference]{IEEEtran}
\IEEEoverridecommandlockouts
\usepackage{cite}
\usepackage{amsmath,amssymb,amsfonts}
\usepackage{algorithmic}
\usepackage{graphicx}
\usepackage{textcomp}
\usepackage{xcolor}
\usepackage{booktabs}
\def\BibTeX{{\rm B\kern-.05em{\sc i\kern-.025em b}\kern-.08em
    T\kern-.1667em\lower.7ex\hbox{E}\kern-.125emX}}
\begin{document}

\title{MALT: Multi-scale Action Learning Transformer for Online Action Detection
\\
\thanks{$^\ast$ Authors with equal contribution.}
\thanks{$^\dagger$ Corresponding author.}
\thanks{This work was supported in part by the National Natural Science Foundation of China under Grant 62372104, in part by the Guangdong Basic and Applied Basic Research Foundation under Grant 2022A1515110518. Besides, we thank the Big Data Computing Center of Southeast University for providing the facility support on the numerical calculations.}
}

\author{
\IEEEauthorblockN{Zhipeng Yang$^{1\ast}$, Ruoyu Wang$^{1\ast}$, Yang Tan$^{1}$, Liping Xie$^{1,2\dagger}$}
    \IEEEauthorblockA{
    1. The Key Laboratory of Measurement and Control of Complex Systems of\\ Engineering, Ministry of Education, School of Automation, Southeast University,\\ Nanjing 210096, China\\
    2. Southeast University Shenzhen Research Institute, Shenzhen 518063, China\\
    E-mail: lpxie@seu.edu.cn
    }
}

\maketitle

\begin{abstract}
Online action detection (OAD) aims to identify ongoing actions from streaming video in real-time, without access to future frames. Since these actions manifest at varying scales of granularity, ranging from coarse to fine, projecting an entire set of action frames to a single latent encoding may result in a lack of local information, necessitating the acquisition of action features across multiple scales. In this paper, we propose a multi-scale action learning transformer (MALT), which includes a novel recurrent decoder (used for feature fusion) that includes fewer parameters and can be trained more efficiently. A hierarchical encoder with multiple encoding branches is further proposed to capture multi-scale action features. The output from the preceding branch is then incrementally input to the subsequent branch as part of a cross-attention calculation. In this way, output features transition from coarse to fine as the branches deepen. We also introduce an explicit frame scoring mechanism employing sparse attention, which filters irrelevant frames more efficiently, without requiring an additional network. The proposed method achieved state-of-the-art performance on two benchmark datasets (THUMOS’14 and TVSeries), outperforming all existing models used for comparison, with an mAP of 0.2\% for THUMOS'14 and an mcAP of 0.1\% for TVseries.
\end{abstract}


\section{Introduction}
Online action detection (OAD)~\cite{oad} has demonstrated considerable potential across a diverse range of fields, including autonomous driving, online anomaly detection, and intelligent surveillance. In contrast to offline methods that assume an entire video is observable, all future frames are unobservable with online techniques. Recent studies~\cite{oadtr,lstr,gatehub,testra,liu2022end,zhang2022actionformer} have utilized transformers~\cite{transformer} to model sequential frames via attention mechanisms. Corresponding encoder units employing cross-attention can also learn attention weights for use in modeling the relationships among input frames, assigning their contributions to the current action prediction. However, these models often rely on a single encoder-decoder architecture~\cite{oadtr,lstr,li2023lighter}, which fails to capture action features across multiple scales or enhance the concentration of some local features~\cite{multioad,twostream}. As single encoders can only model all frames within a global range, the resulting features may lack diversity in time and granularity~\cite{cheng2021coarse}, despite LSTR~\cite{lstr} dividing input sequential frames into short-term and long-term memory. This can be problematic in practical scenarios, as actions often manifest at varying scales of granularity (i.e., the resolution of an entire action), from coarse to fine. In this case, coarse granularity data only provide a rough image of the action while fine granularity offers smaller details. For example, Fig. 2 shows the breaking action in nine-ball, which includes several components, such as a pre-shot routine, cuing of the ball, and striking. As such, projecting the entire set of action frames to one latent encoding may result in a lack of local information. This is because encoder units can only learn weights for all action frames and not for each component individually, producing only a rough picture of the breaking action. It is therefore crucial to capture action features across multiple granularity scales. Irrelevant frames pose another challenge for OAD, as not every historical frame is informative for current frame prediction. This can lead to incorrect projections and large time complexity while encoding entire frame sets. GateHUB~\cite{gatehub} addressed this issue by introducing a GHU to enhance informative history, though the inclusion of a scoring network increased the number of required parameters.

\begin{figure}[t]
	\centering
	\includegraphics[width=10cm]{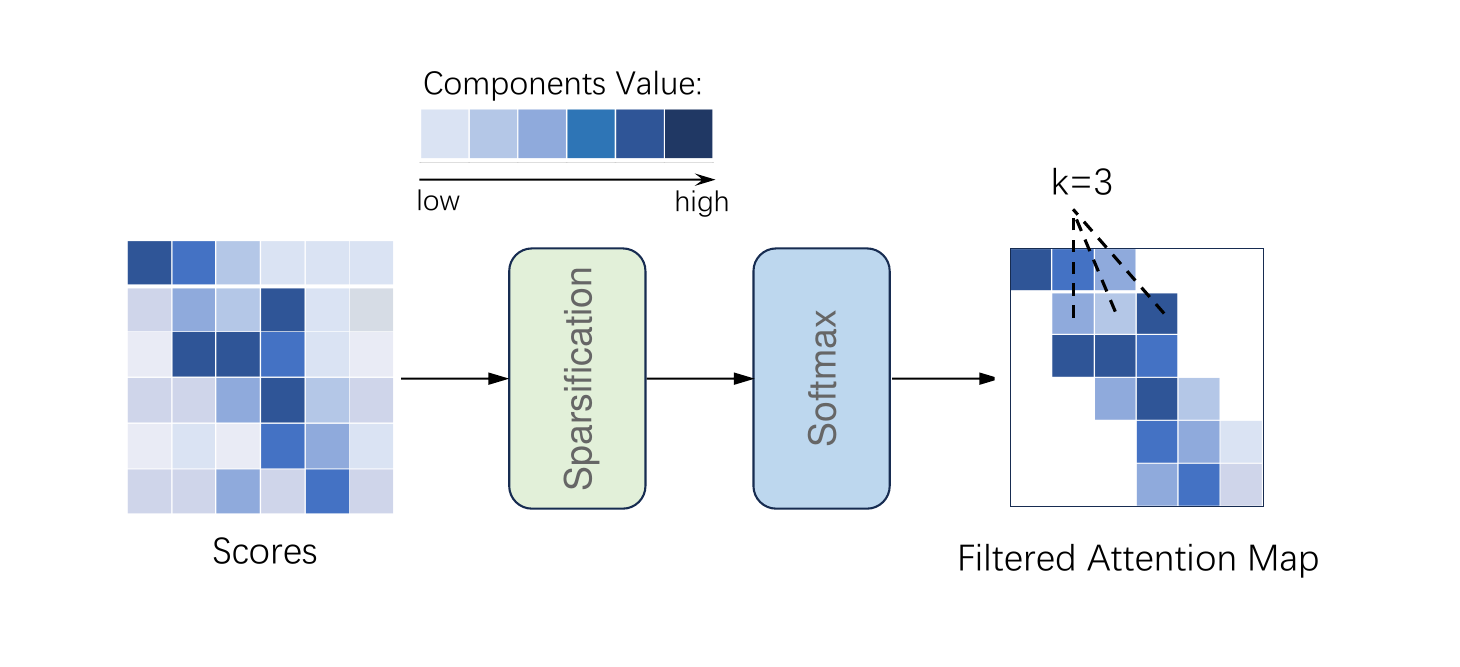}
	\caption{\textbf{An illustration of sparse attention.} A matrix was used to represent the attention map between keys and queries. Sparse attention only selects the top-k largest values from each row, which can be used to explicitly score and filter historical frames. For example, $k=3$ was chosen in this figure, such that only the top-3 largest values were selected in each row.}
	\label{chutian}
\end{figure}

\begin{figure*}[htbp]
	\centering
	\includegraphics[width=18cm]{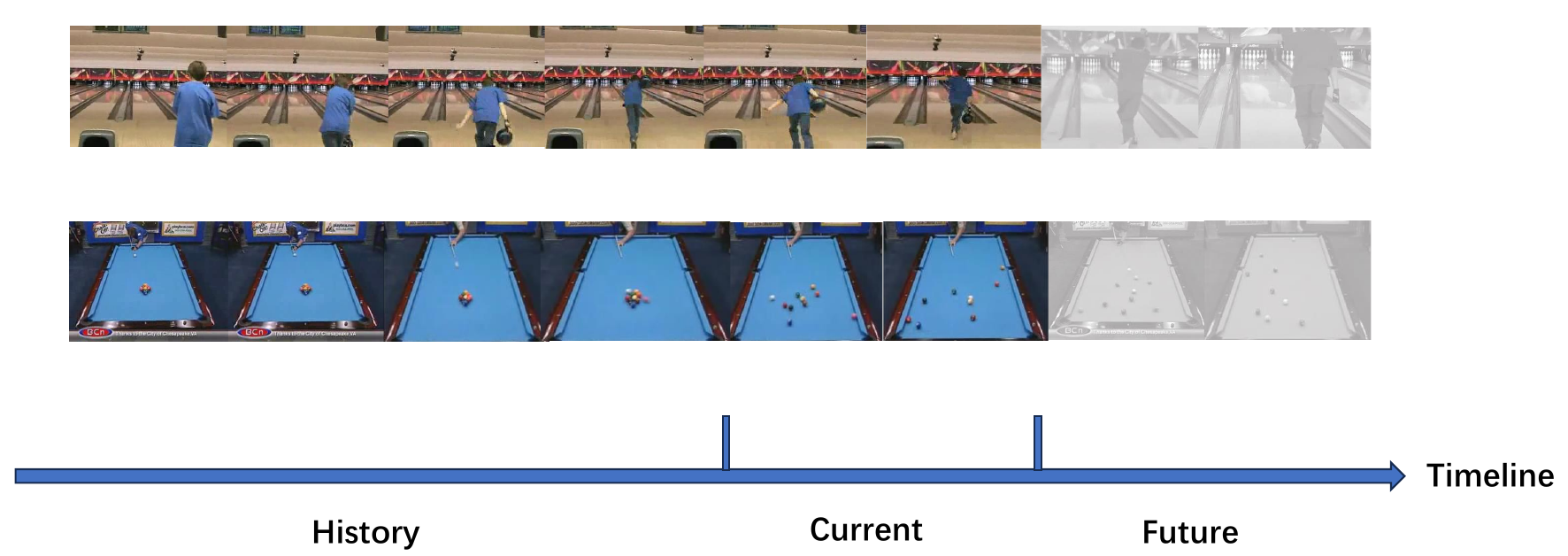}
	\caption{\textbf{An example of OAD.} Nine-ball includes a breaking action that consists of a pre-shot routine, cuing of the ball, and striking. Projecting the entire set of action frames to a single latent encoding may lead to a lack of local information, since encoder units can only learn weights for all action frames and not for each specific component, producing only a rough image of the breaking action.}
	\label{chutian}
\end{figure*}

To address these challenges, we propose a novel multi-scale action learning transformer (MALT) that employs a multi-encoder to single-decoder architecture. A hierarchical encoder was utilized to capture multi-scale action features and a recurrent decoder was employed for feature fusion. Specifically, the encoder includes multiple encoding branches of varying depths, each of which requires the entire frame history as input, to output a latent action feature representation. In each branch, sparse cross-attention is first employed to compress historical frames into a fixed-length latent vector and filter irrelevant frames. Preceding branches are then used to enhance the features from subsequent branches. In this process, coarse granularity features from preceding branches are used to calculate increasingly fine-grained features from the input vectors. More specifically, the corresponding output is incrementally fed into the subsequent branch as part of the input used to calculate cross-attention. In this way, as the branches deepen, output features transition from coarse to fine-grained representations. The hierarchical encoder then outputs an action feature sequence, including a series of features at different granularity scales. A recurrent decoder was also utilized to efficiently fuse multiple action features in a stage-by-stage process, in which the outputs from a previous stage were iteratively provided as inputs to the next stage. As a result, the recurrent decoder includes fewer parameters and can be trained more efficiently. In this step, multiple action features are calculated and used to acquire the outputs.

\begin{figure*}[htbp]
	\centering
	\includegraphics[width=18cm]{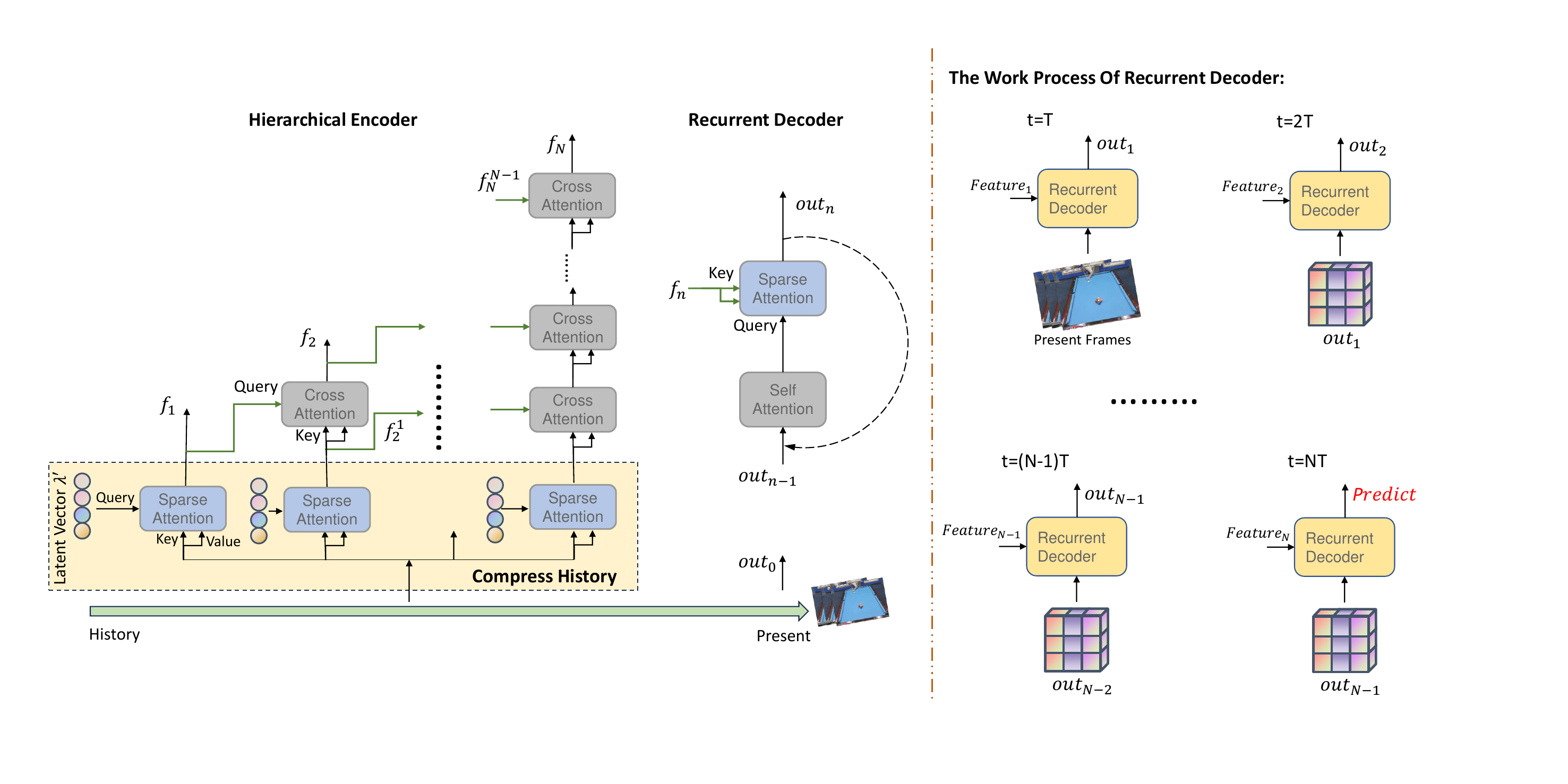}
	\caption{\textbf{An overview of MALT.} The hierarchical encoder structure composed of multiple composite encoding branches, used to calculate the cross-attention between other branches. This process encodes historical frames $M_L$ into an action feature sequence $\{g_1,g_2,...,g_N\}$, including a series at different scales. A recurrent decoder then fuses multiple features in a stage-by-stage fashion.}
	\label{chutian}
\end{figure*}

We also introduce an explicit frame scoring mechanism employing sparse attention~\cite{sparse,ke2018sparse}, which evaluates historical frames based on how informative they are for current frame prediction. In this process, the historical frame attention map is implemented as a scoring mask to select the top-k contributing elements. This approach explicitly filters irrelevant frames without an additional network and thus achieves higher accuracy with less runtime. The proposed MALT network was verified using two common benchmark datasets: THUMOS’14~\cite{thumos} and TVSeries~\cite{oad}. A series of ablation experiments demonstrated the effectiveness of our proposed method, which achieved state-of-the-art results and outperformed all existing models used for comparison, with an mAP of 0.2\% for THUMOS'14 and an mcAP of 0.1\% for TVseries. The primary contributions of this study can be summarized as follows:
\begin{itemize}
    \item We propose the MALT network, used to capture action features across multiple scales. MALT is comprised of a hierarchical encoder (used to encode historical frames into a coarse-to-fine action feature sequence) and a recurrent decoder (employed for efficient feature fusion).
    \item We introduce an explicit frame scoring mechanism employing sparse attention. This approach explicitly filters irrelevant frames without the need for an additional network.
    \item A series of ablation experiments demonstrated the effectiveness of our proposed method, which achieved state-of-the-art results and outperformed all existing models used for comparison, with an mAP of 0.2\% for THUMOS'14 and an mcAP of 0.1\% for TVseries.
\end{itemize}

\section{RELATED WORK}
A. \emph{Online Action Detection}

OAD~\cite{oad} relies entirely on historical and present data to identify actions within a video stream, lacking any foreknowledge of subsequent frames. This task was first introduced by De Geest \textit{et al.}~\cite{oad} who developed the TVseries dataset and implemented several different techniques. IDN~\cite{eun2020learning} learns discriminative features and accumulates only relevant information for present samples. LAP-Net~\cite{qu2020lap} introduces an adaptive sampling strategy to obtain optimal features, while PKD~\cite{zhao2020privileged} employs curriculum learning to transfer knowledge from offline models to OAD tasks. A popular theory suggests that anticipating potential future information could enhance the decision-making process for a current frame. For instance, reinforced encoder-decoder (RED) networks~\cite{red} utilizes LSTM~\cite{hochreiter1997long} to anticipate future events using historical information. Wang et al. proposed OadTR~\cite{oadtr}, which applies a transformer~\cite{transformer} to the prediction of future information by performing action anticipation. However, the techniques used to assist in action detection (by anticipating future events) often fail to fully utilize information from historical frames.

Since online action detection only accesses historical frames, the processing of previous samples poses a key challenge in OAD problems. TRN~\cite{xu2019temporal} utilizes hidden states in LSTM~\cite{hochreiter1997long} to retain a complete history during inference, though it encounters difficulties in effectively modeling long temporal dependencies. Colar~\cite{colar} designed a two-branch system in which a static branch compared a frame with representative samples, delivering complementary information to a dynamic branch. LSTR~\cite{lstr} explicitly divides the entire history into long- and short-term memories using an encoder-decoder architecture for model construction. However, this approach fails to consider that not every historical frame carries useful information. GateHub~\cite{gatehub} introduced a gating mechanism used to filter out redundant details and noise in historical sequences. Another transformer-based architecture, TesTra~\cite{testra}, addressed the growing computational complexity that resulted from an increasing number of frames by employing two types of temporal smoothing kernels. In this paper, we fuse features on different time scales and apply sparse-attention mechanisms to filter invalid frames, thereby modeling long-duration histories more intuitively.

B. \emph{Feature Fusion}

This section considers combinations of multi-scale action features and briefly reviews common fusion strategies. Some dual-stream networks have been developed from these techniques, in which multimodal features are fused prior to the input model \cite{b1,b2,b3}. Other studies have performed a similar feature fusion step in the last layer of the network \cite{b4,b5}. Feature pyramid networks~\cite{lin2017feature} combine pyramid levels using upsampling and addition in a top-down approach, which has become a standard technique for processing multi-scale features. CBNet~\cite{liu2020cbnet} feeds output features from previous levels as part of the input features provided to the subsequent level in a stage-by-stage strategy. Wu et al.~\cite{moniruzzaman2021human} proposed a novel multi-modal dual-stream 3D network structure that can learn spatio-temporal features from depth residual dynamic image sequences and pose estimation map sequences. Wang et al.~\cite{wu2021spatiotemporal} proposed a new lifelong multi-view subspace learning framework, used to identify potential relationships among different views. MGSE~\cite{zhou2023multidimensional} can fuse local and global spatial features for action recognition in videos. However, this approach treats each frame as a two-dimensional vector and simultaneously considers information from the entire video, which is not applicable to our specific task. To our knowledge, the present study represents the first attempt to utilize a feature fusion strategy for online action detection.

\section{METHODOLOGY}
\subsection{Problem Definition}

Given a streaming video sequence $\boldsymbol{h}=[h_t]_{t=-T}^0$, our task is to classify ongoing actions $y_0 \in \{ 0, 1,..., C\}$ occurring at the current frame $h_0$, where 
$C$ is the total number of actions. Since future frames $\{ h_1, h_2,... \}$ are unavailable, this prediction is based entirely on historical and current frames. 

\subsection{Sparse Attention}

The relationships among historical frames in the input video sequence can be modeled by leveraging transformer attention mechanisms~\cite{transformer}, which enables the acquisition of an attention map representing the level of correlation between historical frames. In the case of the vanilla cross-attention mechanism, given the input sequences $X_1\in \mathbf{R}^{L_1\times D}$ and $X_2\in \mathbf{R}^{L_2\times D}$, the query, key, value, and attention map can be expressed as: 
\begin{equation}
    Q=X_1W_q,
\end{equation} 
\begin{equation}
    K=X_2W_k, 
\end{equation}
\begin{equation}
    V=X_2W_v,
\end{equation}
\begin{equation}
    A=\frac{QK^T}{\sqrt{D}},
\end{equation}
where $L_1$ and $L_2$ denote the number of frames and $D$ represents the channel dimension. In this case, $L_1$ is typically smaller than $L_2$, allowing historical frames to be compressed into a fixed-length vector. However, vanilla cross-attention assigns credit to all components of each frame, some of which may be lacking useful information. 

As such, we can employ a sparse attention mechanism, a modified version of vanilla attention~\cite{sparse}, to explicitly score and filter historical frames (see Fig. 1). Unlike vanilla attention, sparse attention only learns weights for informative frames, while no credit is assigned to irrelevant frames. Vanilla attention then degenerates to sparse attention through a top-k selection. These attention map results ($A$) can be considered scores representing the relevance between two frames, where a lower score indicates a less relevant frame in the current prediction step. Therefore, only the elements with the highest scores need to be retained. In this case, $\emph{k}$ elements are retained for each row using a sparse attention mechanism, the value of which should be smaller than the frame dimensions (a value of $\emph{k}=$370 is selected from empirical results). A sparse-attention mask is then used to select contributing elements in $A$. Specifically, we only select the top-k largest elements from each row in $A$, and retain their values. Other elements are assigned to negative infinity as follows:
\begin{equation}
    \Psi\left(A,k\right)_{ij} =
    \begin{cases}
        A_{ij}\qquad if\quad A_{ij}\geq t_i,   \\
        -\infty \qquad if\quad A_{ij}\leq t_i,  \\
    \end{cases} 
\end{equation}
where $k$ is a hyperparameter that determines the number of retained elements and $t_i$ is the $k^{th}$ largest value in row \emph{i}. The values of irrelevant elements are set to negative infinity and a normalization step is then used to set the probabilities of irrelevant frames to $0$. In this way, the most highly contributing frames are reserved for the attention map, while other irrelevant information is removed. Finally, in the case of the filtered attention map, results from the sparse-attention mechanism are given by:

\begin{equation}     SparseAttn=Softmax\left(\Psi\left(A,k\right)\right)V.
\end{equation}

\subsection{Hierarchical Encoder}

Following the long short-term partition used in LSTR~\cite{lstr}, $\boldsymbol{h}$ is divided into short-term ($M_S$) and long-term ($M_L$) memory components, represented by:
\begin{equation}
M_S=\{ h_{-m_s+1}, h_{-m_s+2},...,h_0 \},    
\end{equation}
\begin{equation}
M_L=\{ h_{-m_l-m_s+1},h_{-m_l-m_s+2},...,h_{-m_s} \},    
\end{equation}
where $m_s \ll m_l$. This hierarchical encoder aims to encode long-term memory $M_L$ into an action feature sequence, including a series of terms at different granularity scales. Previous studies~\cite{jaegle2021perceiver,lstr} have used cross-attention to project the variable length history to a fixed-length learned latent encoding, which can be used for decoding useful temporal contexts. However, as shown in Fig. 2, these single encoder-decoder architectures fail to encode historical frames at varying levels of granularity or enhance the concentration of encoder units for some informative local features~\cite{multioad,twostream}. We therefore introduce a hierarchical encoder composed of multiple encoding branches with varying depths. Each branch requires the full set of historical frames $M_L$ as input, thereby outputting a latent action feature representation $f_n$. This action feature sequence can be represented as $\{f_1,f_2,...,f_N\}$. 

Specifically, the hierarchical encoder consists of $N$ encoding branches with varying depths, where the $n^{th}$ branch comprises $P_n$ stages. As the value of $n$ increases, these output features exhibit increasingly fine-grained granularity. Thus, we selected $P_n=n$, such that the $n^{th}$ branch includes additional stages as $n$ increases. However, directly encoding the historical frames $M_L$ can result in large time complexity when $m_l$ is a large number. Therefore, we first employ sparse cross-attention to compress $M_L \in \mathbf{R}^{m_l \times D}$ into a fixed-length latent vector, thereby filtering irrelevant frames. We also define a learnable latent query vector $\lambda \in \mathbf{R}^{L_n \times D}$, where $L_n$ and $D$ denote the compressed length and the embedding dimension, given $L_n=\frac{L}{2^{n-1}}$ ($L \ll m_l$). Self attention is first applied to $\lambda$ and the output $\lambda'$ is then used as a query in sparse cross attention. The input $M_L$ serves as a key and a value, expressed as:
\begin{equation}
f_n^1=SparseAttn(\lambda',M_L),    
\end{equation}
where $f_n^1 \in \mathbf{R}^{L_n \times D}$ is the output of the first stage for the $n^{th}$ branch. In sparse attention, historical frames of length $m_l$ are compressed to $L_n$, which can be a very small number ($L=32$ in this case). As the encoding branches deepen, $L_n$ becomes smaller and the historical frames are compressed into shorter feature vectors.

Unlike existing techniques~\cite{lstr,gatehub,testra} that simply stack multi-layer cross attention, we employ $n-1$ encoding branches to enhance the features of the $n^{th}$ branch, by iteratively feeding output features from the previous branch as part of the input to subsequent branches. Cross attention was then applied to fuse output features from a given branch stage. As shown in Fig. 3, these previous branch outputs ($f_{n-1}^{p-1}$) serve as queries, while outputs from the same stage ($f_{n}^{p-1}$) in the previous branch serve as keys and values. This process can be expressed as:
\begin{equation}
    f_n^p=CrossAttn(f_{n-1}^{p-1},f_n^{p-1}),
\end{equation}
where \emph{n} and \emph{p} denote the $n^{th}$ branch and the $p^{th}$ stage, respectively. Outputs from the last stage represent an action feature vector denoted by $f_n=f_n^{P_n}$. Note that in the case of $P_n=n$, this gives:
\begin{eqnarray}
  f_n=CrossAttn(f_{n-1}^{n-1},f_n^{n-1}), \nonumber \\
=CrossAttn(f_{n-1},f_n^{n-1}),  
\end{eqnarray}
where $f_n$ maintains identical dimensions with $f_{n-1}$, where $f_1 \in \mathbf{R}^{L \times D}$ and $f_n \in \mathbf{R}^{L \times D}$.

\subsection{Recurrent Decoder}

The hierarchical encoder includes an action feature sequence $\{f_1,f_2,...,f_N\}$ and it is thus crucial to fuse these multi-scale features efficiently. One approach~\cite{lstr} involves using cascaded multi-attention layers and continuously feeding the vector $f_n$ to calculate the cross-attention between $f_n$ and the output from the preceding attention layer. However, cascading layers can lead to overfitting and introduce a large number of training parameters. As such, we develop a recurrent decoder that fuses features in an iterative manner. This decoder works in a stage-by-stage process, in which the outputs from a previous stage are iteratively fed as inputs to the same layer in the next stage, while the vector $f_n$ is used to calculate the outputs. More specifically, we define a vector $Q\in \mathbf{R}^{L \times D}$ for use in this decoding operation. Here, Q is sampled from the short term memory $\{ h_{-m_s+1}, h_{-m_s+2},...,h_0 \}$ in the first stage, while the outputs $Out_{n-1}$ in the ${n-1}^{th}$ stage serve as Q for the subsequent $n^{th}$ stage. Unlike the previous step using cascading multi-attention layers, the same attention layers are used at each stage, which is equivalent to sharing weights across multi-attention layers~\cite{takase2021lessons,han2021connection}. As a result, the recurrent decoder has fewer parameters and can be trained more efficiently. The presented decoding operation includes two steps. In the $n^{th}$ stage, self-attention is first applied to $Q$ and sparse cross-attention is then calculated using the feature vector $f_n$. The outputs from this self-attention step ($Q'$) are then used as queries while the input feature vector $f_n$ serves as a set of keys and values. The outputs $Out_n$ then serve as $Q$ in the subsequent (${n+1}^{th})$ decoding stage. Since the number of action features is represented by $N$, this produces a total of $N$ decoding stages. Finally, outputs from the last decoding stage are fed into the classifier layer and used for prediction.

\subsection{MALT Training}
Previous studies~\cite{lstr} have applied cross-entropy loss to all present frames as follows:
\begin{equation}
    \mathcal{L}=-\sum_{i=t-M_S+1}^t \sum_{j=1}^C \mathbf{y}_{i,j} log\mathbf{\hat{y}}_{i,j},
\end{equation}
where $y_{i,j}$ is the ground truth action label and $\hat{y}_{i,j}$ is the prediction from the output token corresponding to time \emph{t}. In this case, we apply the classifier layer and the loss function to the action feature sequence $\{f_1,f_2,...,f_N\}$ and add an auxiliary loss term given by~\cite{lee2015deeply,wang2015training}:
\begin{equation}
    \mathcal{L}'=\alpha \mathcal{L}+ \sum_{n=1}^N \beta_n \mathcal{L}_n
\end{equation}
to the final overall loss, where $\alpha$ and $\beta_n$ are constants used to balance the loss types.

\section{EXPERIMENTS}
\subsection{Datasets}
The proposed MALT network was verified across two OAD benchmark datasets: TVSeries~\cite{oad} and THUMOS’14~\cite{thumos}.

\emph{THUMOS'14:} This dataset consists of 413 untrimmed videos annotated with 20 actions. Following the test protocols for existing OAD methods~\cite{xu2019temporal,eun2020learning}, we allocated 200 videos as the training set and 213 videos as the test set.

\emph{TVSeries:} This dataset includes 6 popular TV shows, with a total of 27 episodes spanning over 16 hours. The data are labeled with 30 everyday actions occurring in real-world scenarios, such as "open a door," "throw a punch," and "drink."

\subsection{Implementation Details}
The feature encoding step adopted two-stream deep features~\cite{xiong2016cuhk} pre-trained on Kinetics, following the method proposed by Xu et al.~\cite{lstr}. More specifically, features were extracted from the central frames of each section of the visual model using a ResNet-50~\cite{he2016deep} architecture. The BN-Inception~\cite{ioffe2015batch} framework was then employed in the motion model. The proposed MALT system was implemented in PyTorch and an Nvidia 3090 GPU was employed to perform all experiments. The transformer design included 16 multi-attention heads and hidden dimensions of 1024. Model weights were trained using the Adam~\cite{kingma2014adam} optimizer with a learning rate of $5 \times 10^{-5}$ and 25 epochs. Results from the encoder were immediately derived when calculating the loss function and used to obtain auxiliary losses, which then underwent back-propagation along with the final model losses. Evaluation protocols included the use of mean average precision (mAP) to evaluate performance on THUMOS'14. The calibrated average precision (cAP)~\cite{oad} was also applied to the TVSeries results, thereby balancing positive and negative samples.

\subsection{Quantitative Comparisons}
\begin{table}[!htbp] 
\begin{center}
\begin{tabular}{ccc}
\toprule 
\textbf{Method} & \textbf{THUMOS’14  (mAP)} & \textbf{TVSeries  (mcAP)}\\
\midrule
TRN~\cite{xu2019temporal} & 62.1 & 86.2 \\
IDN~\cite{eun2020learning} & 60.3 & 86.1 \\
PKD~\cite{zhao2020privileged} &   64.5   &    86.4   \\
OadTR~\cite{oadtr} & 65.2 & 87.2 \\
Colar~\cite{colar} & 66.9 & 88.1 \\
WOAD~\cite{gao2021woad} & 67.1 & -\\
LSTR~\cite{lstr} & 69.5 & 89.1  \\
GateHUB~\cite{gatehub} & 70.7 & 89.6 \\
TeSTra~\cite{testra} & 71.2 & -\\
FCRL~\cite{leng2023online} & 68.3 & 88.4 \\
Ours & 71.4 & 89.7 \\
\bottomrule
\end{tabular}
\end{center}

TABLE \uppercase\expandafter{\romannumeral1}: A comparison of two-stream network feature extractors pre-trained on Kinetics-400 using THOUMOS’14 and TVSeries.
\end{table}

Table 1 compares the proposed MALT network with existing state-of-the-art methods applied to the THUMOS’14~\cite{thumos} and TVSeries~\cite{oad} datasets, used to verify the effectiveness of our model. Specifically, compared to LSTR~\cite{lstr}, which adopts transformer encoder units to model historical frames without the use of filtering frames, MALT produced a 1.9\% mAP increase for THUMOS'14 and a 0.6\% mcAP gain for TVseries. Furthermore, enhancing the features in historical frames by fusing multi-scale features in MALT led to a 0.7\% mAP increase for THUMOS’14 (compared to GateHUB~\cite{gatehub}, which adopts a future-augmented history). It was evident that MALT outperformed all existing models used for comparison, with an mAP of 0.2\% for THUMOS'14 and an mcAP of 0.1\% for TVseries, which confirms that our method achieved promising OAD results.

\subsection{Ablation Studies}
A series of ablation experiments were performed to evaluate the effects of our proposed sparse-attention mechanism, aux-loss term, and proposed decoder architecture used to fuse multi-scale features. Extensive tests were performed using the THUMOS'14 dataset and all models were trained from scratch. The efficacy of each component is illustrated in Table II.

\textbf{Impact of Sparse Attention}
We first verified the role of sparse attention in MALT. Table \uppercase\expandafter{\romannumeral2} demonstrates that including a sparse attention mechanism improved performance by 0.2\%. The results shown in Fig. 1 indicate the inclusion of this mechanism reduced computational complexity and alleviated the impact of irrelevant frames on action inference. Since the hierarchical encoder was used to extract multi-scale action features, the gain produced by sparsity during each attention operation increased exponentially as the encoder layers deepened.  

\textbf{MALT Recurrent-decoder}
The recurrent decoder included in MALT sequentially queries the multi-scale action features extracted by the hierarchical encoder, rather than encoding historical frames only once and inputting them to the decoder. As such, we suggest our architecture facilitates a more comprehensive consideration of different time-scale action representations compared to single scale modeling. The performance gap (71.4 to 69.5 in HICO-DET) illustrated in Table \uppercase\expandafter{\romannumeral2} indicates the recurrent decoder alleviated any issues with single-scale modeling.

\textbf{Impact of Aux-loss}
As the MALT encoder exhibits multiple layers, we applied an auxiliary detection head to stabilize the convergence of deeper encoder layers. We directly predicted the actions of encoder outputs and calculated the resulting cross entropy loss, which was then added to the final output loss function with an included weighting term. Finally, we set this weight value to 0.4 after fine tuning during the training process. It is evident from Table \uppercase\expandafter{\romannumeral2} that a 0.6\% decrease in mAP occurred when auxiliary loss was excluded, confirming that aux-loss enhanced performance.

\begin{table}[!ht]
\begin{center}
\begin{tabular}{cc}
\toprule
\textbf{Method} & \textbf{THUMOS’14 / mAP (\%)} \\
\midrule
MALT & 71.4  \\
w/o Sparse-attention & 71.2  \\
w/o Recurrent-decoder & 69.5  \\
w/o Aux-loss & 70.8 \\
\bottomrule 
\end{tabular}
\end{center}

TABLE \uppercase\expandafter{\romannumeral2}: A comparison of MALT with the baseline LSTR applied to the THUMOS'14 test set.
\end{table}

\subsection{Design Choices}

Individual components were optimized by investigating the influence of various hyperparameters on model performance, as described below.

\begin{table}[!htbp]
\centering
\begin{center}
\begin{tabular}{cc}
  \hline
  \textbf{Encoder Branches} & \textbf{mAP (\%)}
  \\
  \hline
   1 & 70.9\\
   2 & 71.4\\
   3 & 71.3\\
   4 & 71.0\\
  \hline
\end{tabular}
\end{center}
{(a)}
\end{table}
\begin{table}[!htbp]
\begin{center}
\begin{tabular}{cc}
  \hline
  \textbf{Value of k} &  \textbf{mAP (\%)}
  \\
  \hline
   290 & 69.6\\
   310 & 70.7\\
   350 & 71.3\\
   370 & 71.4\\
   390 & 71.4\\
   410 & 71.3\\
   430 & 71.3\\
   450 & 71.2\\
  \hline
\end{tabular}
\end{center}
\quad \quad \quad \quad \quad \quad \quad \quad \quad \quad \quad\quad\quad \quad\quad{(b)}

TABLE \uppercase\expandafter{\romannumeral3}: An evaluation of design choices for our proposed technique, including (a) various branch quantities in the hierarchical encoder and (b) k values in the sparse attention mechanism.
\end{table}

\textbf{Encoder Branches:} We first analyzed the effects of various encoder branch quantities, which suggested that additional encoder branches allowed the hierarchical encoder to model historical frames across a more diverse range of granularity scales. However, increasing the number of encoder branches can also lead to increased time complexity and overfitting. As such, various quantities were tested, as shown in Table \uppercase\expandafter{\romannumeral3}(a). Our method achieved the best results for the case of N = 2, while the time complexity remained lower than that of other models.

\textbf{Value of k:} We also selected the degree of historical frame filtering by adjusting the k value. Following LSTR protocols~\cite{lstr}, we chose a length of 512 seconds for the historical frames. When compressing these frames into a fixed-length vector, we both evaluated historical frames and filtered irrelevant frames using a top-k selection. In this case, a smaller value for k resulted in less historical information being retained. Since the performance of k\textgreater450 was the same as that without sparse attention, we set 450 as the upper bound. Table \uppercase\expandafter{\romannumeral3}(b) demonstrates that the utilization efficiency for historical frames was highest for k=370 or 390.

\begin{table}[!ht]
\centering
\begin{center}
\begin{tabular}{lc}
  \hline
  \textbf{Feature Fusion} & \textbf{mAP (\%)}
  \\
  \hline
Add & 69.8  \\
Cascaded Layers & 70.8  \\
Recurrent-Decoder & 71.4  \\
  \hline
\end{tabular}
\end{center}
{(a)}
\end{table}
\begin{table}[!ht]
\begin{center}
\begin{tabular}{lc}
  \hline
  \textbf{Filtering Frames} &  \textbf{mAP (\%)}
  \\
  \hline
Not Filtered & 71.2  \\
GHU~\cite{gatehub} & 71.2  \\
Sparse Attention~\cite{sparse} & 71.4  \\
  \hline
\end{tabular}
\end{center}
\quad \quad \quad \quad \quad \quad \quad \quad \quad \quad \quad\quad\quad \quad\quad{(b)}

TABLE \uppercase\expandafter{\romannumeral4}: Design choices for the proposed technique, including methods used for (a) feature fusion and (b) filtering irrelevant frames.
\end{table}

\textbf{Feature Fusion:} We also compared different methods used to fuse multiple action features by first simply adding features and then feeding the results into the decoder. We cascaded the multi-attention layers without sharing the weights and continuously input the resulting features used to calculate cross-attention. As indicated by Table \uppercase\expandafter{\romannumeral4}(a), our proposed recurrent decoder was the best choice for feature fusion.

\textbf{Filtering Irrelevant Frames:} Our frame filtering method was also compared with a GHU~\cite{gatehub}, which employs a scoring network to determine how informative historical frames are to current frame prediction tasks. This approach explicitly filters irrelevant frames without the need for an additional network. Table \uppercase\expandafter{\romannumeral4}(b) demonstrates that our method achieved the best performance.

\section{CONCLUSION}
This paper proposed a multi-scale action learning transformer (MALT), composed of a hierarchical encoder (used to capture multi-scale action features) and a recurrent decoder (used for feature fusion). The recurrent decoder includes fewer parameters and can therefore be trained more efficiently. The hierarchical encoder exhibits multiple encoding branches of varying depths. The output from the preceding branch was incrementally fed into subsequent branches as part of the input. In this way, output features transitioned from coarse to fine-grained as the branches deepened. We also introduced an explicit frame scoring mechanism that employed sparse attention, which could filter irrelevant frames more efficiently without requiring an additional network. MALT outperformed all existing models used for comparison, with an mAP of 0.2\% for THUMOS'14 and an mcAP of 0.1\% for TVseries. However, there are also some limitations of MALT, which should be discussed in the future works. For example, as the number of encoder branches increases, historical frames can be encoded on more scales, which could also lead to increased time complexity and overfitting.

\bibliographystyle{unsrt}
\bibliography{reference}

\end{document}